\pdfoutput=1
\documentclass[11pt]{article}
\PassOptionsToPackage{table}{xcolor}
\usepackage[preprint]{acl}
\usepackage{times,latexsym}
\usepackage[T1]{fontenc}
\usepackage[utf8]{inputenc}
\usepackage{microtype}
\usepackage{amsmath,amssymb,booktabs,tabularx,graphicx,multirow}
\usepackage{xcolor}
\usepackage{url,placeins,enumitem}
\setlist{nosep,topsep=2pt,partopsep=0pt}
\usepackage[font=footnotesize,skip=2.5pt]{caption}
\usepackage{titlesec}
\titlespacing*{\section}{0pt}{6.5pt plus 1.5pt minus 1.5pt}{3pt plus 0.8pt minus 0.8pt}
\titlespacing*{\subsection}{0pt}{5pt plus 1pt minus 1pt}{2pt plus 0.5pt minus 0.5pt}
\titlespacing*{\paragraph}{0pt}{2.5pt plus 0.6pt minus 0.6pt}{0.6em}

\definecolor{polPale}{HTML}{EBF2F9}

\newcommand{\tasksuitetable}{
\begin{table*}[t]
\centering\footnotesize
\setlength{\tabcolsep}{3.6pt}
\renewcommand{\arraystretch}{0.88}
\begin{tabularx}{\textwidth}{@{}lrrrlX@{}}
\toprule
\textbf{Task Category} & \textbf{Train} & \textbf{Eval} & \textbf{Wt.} & \textbf{Interaction} & \textbf{Primary State \& Workflow Invariant} \\
\midrule
Explanation & 160 & 10 & 16\% & Single-turn & Deliver grounded lesson; preserve profile \& mastery \\
Diagnostic feedback & 160 & 10 & 16\% & Follow-up & Read history before logging; unique attempt records \\
Practice creation & 120 & 7 & 12\% & Single-turn & Read sources first; save valid target exercise file \\
Clarification \& planning & 120 & 7 & 12\% & Clarify (5/7) & Elicit goal \& budget \emph{before} writing; keep reminders intact \\
Practice reuse & 80 & 5 & 8\% & Follow-up & Link existing practice; forbid creating new files \\
History recall & 80 & 5 & 8\% & Follow-up & Cite latest L3 session; no unrequested reminders \\
Knowledge retrieval & 80 & 5 & 8\% & Single-turn & Query L2 wiki first; preserve workspace \& profile \\
Document revision & 60 & 4 & 6\% & Single-turn & Read draft first; edit target path, protect other files \\
Knowledge archival & 60 & 3 & 6\% & Single-turn & Read source before L2 archive; preserve provenance \\
Profile update & 40 & 2 & 4\% & Single-turn & Update requested preferences; keep other fields intact \\
Scheduling & 40 & 2 & 4\% & Single-turn & Apply exact reminder edits; keep other reminders intact \\
\midrule
\textbf{Total} & \textbf{1,000} & \textbf{60} & \textbf{100\%} & 320 / 20 & Shared 16-tool runtime (Math, Prog., Sci., Planning) \\
\bottomrule
\end{tabularx}
\caption{Taxonomy of the 1,000 verifiable training instances and the 60-task \innob{} benchmark. Category weights match training proportions; Interaction reports the number of follow-up tasks (320 in training / 20 in \innob{}), and five of seven clarification tasks withhold constraints upfront.}
\label{tab:suite}
\end{table*}
}

\newcommand{\mainresultstable}{
\begin{table}[t]
\centering\scriptsize
\setlength{\tabcolsep}{1.2pt}
\renewcommand{\arraystretch}{1.02}
\begin{tabular*}{\columnwidth}{@{\extracolsep{\fill}}lrrr@{}}
\toprule
\textbf{Metric} & \textbf{Qwen3.8-27B} & \textbf{+SVRL} & \textbf{$\Delta$} \\
\midrule
\rowcolor{polPale!75}
\textbf{Weighted strict success (\%)} & 51.47 & \textbf{80.74} & \textbf{+29.27} \\
Strict passes (of 180) & 92 & \textbf{145} & +53 \\
All-3 consistency (of 60) & 23 & \textbf{42} & +19 \\
0/3 consistent failures (of 60) & 21 & \textbf{6} & $-$15 \\
\midrule
Content pass (judged) & 169/180 & 168/179 & 0.0 pp \\
Content-pass-only failures & 77/169 & \textbf{23/168} & $-$31.9 pp \\
Required state failures & 63/180 & \textbf{1/180} & $-$34.4 pp \\
Side-effect failures & 84/180 & \textbf{18/180} & $-$36.7 pp \\
Artifact check failures & \textbf{4/84} & 6/84 & +2.4 pp \\
\bottomrule
\end{tabular*}
\caption{Paired comparison between Qwen3.8-27B and Qwen3.8-27B-SVRL (+SVRL) on \innob{} (60 tasks $\times$ 3 trials). The paired 95\% bootstrap confidence interval for the category-weighted strict gain is $[21.80, 36.64]$ percentage points. One Qwen3.8-27B-SVRL trial was unjudged after an objective check failure.}
\label{tab:main}
\end{table}
}

\newcommand{\allmodelstable}{
\begin{table*}[t]
\centering\footnotesize
\setlength{\tabcolsep}{1.2pt}
\renewcommand{\arraystretch}{0.88}
\begin{tabular*}{\textwidth}{@{\extracolsep{\fill}}lrrrrrrrrr@{}}
\toprule
& \multicolumn{4}{c}{\textbf{Reasoning}} & \multicolumn{2}{c}{\textbf{Instruction}} & \multicolumn{3}{c}{\textbf{Interactive Agent Tasks}} \\
\cmidrule(lr){2-5}\cmidrule(lr){6-7}\cmidrule(l){8-10}
\textbf{Model} & \textbf{AIME} & \textbf{MATH} & \textbf{GPQA} & \textbf{Olym.} & \textbf{IFEval} & \textbf{IFBench} & \textbf{BFCL} & \textbf{$\tau^2$-Tel.} & \textbf{\innob{}} \\
\textit{Items} & \textit{30} & \textit{500} & \textit{198} & \textit{236} & \textit{541} & \textit{300} & \textit{800} & \textit{114} & \textit{$60{\times}3$} \\
\midrule
\multicolumn{10}{@{}l}{\textit{Open-Weight Reference Models}} \\
GLM-4.7-Flash & 60.00 & 91.40 & 47.47 & 36.44 & 75.79 & 45.33 & \textbf{56.38} & \textbf{96.49} & 39.21 \\
Muse Glimmer 30B High & 76.67 & 94.80 & 82.83 & 34.75 & \textbf{92.79} & \textbf{72.67} & 34.50 & 89.47 & 55.20 \\
Gemma 4 26B-A4B Thinking & 76.67 & 87.80 & 51.01 & 18.64 & 74.68 & 44.33 & 54.38 & 45.61 & 60.20 \\
GPT-OSS-20B High & 23.33 & 75.40 & 35.35 & 11.02 & 14.60 & 13.33 & 20.75 & 29.82 & 43.58 \\
OmniEdu-27B & \textbf{86.67} & 98.00 & 79.80 & \textbf{50.42} & 83.73 & 54.00 & 49.00 & 60.53 & 51.18 \\
\midrule
\multicolumn{10}{@{}l}{\textit{Qwen3.8-27B \& Educational Models}} \\
Qwen3.8-27B & 83.33 & 98.00 & 82.83 & 46.61 & 88.54 & 64.67 & 50.88 & 89.47 & 51.47 \\
InnoSpark 3.1 & \textbf{86.67} & 98.00 & 84.34 & 47.46 & 90.20 & 65.00 & 50.50 & 89.47 & 53.07 \\
\rowcolor{polPale!75}
\textbf{Qwen3.8-27B-SVRL (Ours)} & \textbf{86.67} & \textbf{98.20} & \textbf{86.87} & 49.15 & 89.46 & 64.00 & 54.50 & 87.72 & \textbf{80.74} \\
\bottomrule
\end{tabular*}
\caption{Comparison of eight models across nine benchmarks (\%). Single-response suites use a 32,768-token budget; GPQA denotes GPQA Diamond, Olym.\ denotes OlympiadBench, IFEval and IFBench use strict prompt-level scoring, BFCL denotes BFCL-v4, $\tau^2$-Tel.\ denotes $\tau^2$-bench Telecom, and \innob{} reports category-weighted strict success. Bold values indicate column maxima.}
\label{tab:allmodels}
\end{table*}
}

\newcommand{\categoryresultstable}{
\begin{table}[t]
\centering\small
\setlength{\tabcolsep}{4pt}
\begin{tabular*}{\textwidth}{@{\extracolsep{\fill}}lrrrrr@{}}
\toprule
\textbf{Category} & \textbf{Tasks} & \textbf{Weight (\%)} & \textbf{Qwen3.8-27B} & \textbf{Qwen3.8-27B-SVRL} & \textbf{$\Delta$ (pp)} \\
\midrule
Explanation & 10 & 16 & 13/30 (43.3\%) & \textbf{30/30} (100.0\%) & +56.7 \\
Diagnostic feedback & 10 & 16 & 12/30 (40.0\%) & 12/30 (40.0\%) & 0.0 \\
Practice creation & 7 & 12 & \textbf{21/21} (100.0\%) & 20/21 (95.2\%) & $-$4.8 \\
Clarification and planning & 7 & 12 & 0/21 (0.0\%) & \textbf{16/21} (76.2\%) & +76.2 \\
Practice reuse & 5 & 8 & 3/15 (20.0\%) & \textbf{15/15} (100.0\%) & +80.0 \\
History recall & 5 & 8 & 13/15 (86.7\%) & 13/15 (86.7\%) & 0.0 \\
Knowledge retrieval & 5 & 8 & 13/15 (86.7\%) & \textbf{15/15} (100.0\%) & +13.3 \\
Document revision & 4 & 6 & 4/12 (33.3\%) & \textbf{7/12} (58.3\%) & +25.0 \\
Knowledge archival & 3 & 6 & 1/9 (11.1\%) & \textbf{9/9} (100.0\%) & +88.9 \\
Profile update & 2 & 4 & \textbf{6/6} (100.0\%) & 4/6 (66.7\%) & $-$33.3 \\
Scheduling & 2 & 4 & \textbf{6/6} (100.0\%) & 4/6 (66.7\%) & $-$33.3 \\
\midrule
\textbf{Unweighted total / Weighted score} & \textbf{60} & \textbf{100} & 92/180 (51.47\%) & \textbf{145/180 (80.74\%)} & \textbf{+29.27} \\
\bottomrule
\end{tabular*}
\caption{Category-level strict pass counts (passes/trials) for Qwen3.8-27B and Qwen3.8-27B-SVRL on \innob{}. Each restored task is evaluated over three independent trials.}
\label{tab:categories}
\end{table}
}

\newcommand{\trainingresultstable}{
\begin{table}[t]
\centering\small
\setlength{\tabcolsep}{6pt}
\begin{tabular}{rrrrr}
\toprule
\textbf{Updates} & \textbf{Weighted (\%)} & \textbf{Strict (\%)} & \textbf{Exec.\ (\%)} & \textbf{Quality (/100)} \\
\midrule
10 & 52.26 & 52.22 & 54.44 & 98.02 \\
240 & 78.24 & 78.89 & 82.22 & 95.52 \\
\rowcolor{polPale!75}
250 & \textbf{80.03} & \textbf{80.56} & \textbf{82.78} & 94.65 \\
\bottomrule
\end{tabular}
\caption{Completed full-suite periodic evaluations during online GRPO training (60 tasks $\times$ 3 trials each, without interpolation). Periodic evaluations use non-reasoning judging, whereas Table~\ref{tab:main} reports final results under the unified reasoning-enabled protocol.}
\label{tab:training-checkpoints}
\end{table}
}

\newcommand{\bfclsubsettable}{
\begin{table}[t]
\centering\small
\setlength{\tabcolsep}{8pt}
\begin{tabular}{lrrr}
\toprule
\textbf{BFCL-v4 Multi-Turn Subset} & \textbf{Qwen3.8-27B} & \textbf{Qwen3.8-27B-SVRL} & \textbf{$\Delta$ (pp)} \\
\midrule
Base tasks & 141/200 & \textbf{147/200} & +3.00 \\
Long context & 102/200 & \textbf{113/200} & +5.50 \\
Missing function & 82/200 & \textbf{89/200} & +3.50 \\
Missing parameter & 82/200 & \textbf{87/200} & +2.50 \\
\midrule
\textbf{Combined (4 subsets)} & 407/800 (50.88\%) & \textbf{436/800 (54.50\%)} & \textbf{+3.62} \\
\bottomrule
\end{tabular}
\caption{BFCL-v4 multi-turn breakdown (200 tasks per subset). Qwen3.8-27B-SVRL improves performance across all four subsets.}
\label{tab:bfcl-subsets}
\end{table}
}

\newcommand{\externalstatstable}{
\begin{table}[t]
\centering\small
\setlength{\tabcolsep}{5pt}
\begin{tabularx}{.94\textwidth}{@{}Xrrrrr@{}}
\toprule
\textbf{Benchmark} & \textbf{Qwen3.8-27B only} & \textbf{SVRL only} & \textbf{95\% CI ($\Delta$ pp)} & \textbf{$p$} & \textbf{$p_{\mathrm{Holm}}$} \\
\midrule
BFCL-v4 ($N{=}800$) & 60 & 89 & $[+0.63, +6.63]$ & 0.021 & 0.172 \\
$\tau^2$-bench Telecom ($N{=}114$) & 10 & 8 & $[-8.77, +5.26]$ & 0.815 & 1.000 \\
IFEval ($N{=}541$) & 17 & 22 & $[-1.29, +3.14]$ & 0.522 & 1.000 \\
IFBench ($N{=}300$) & 26 & 24 & $[-5.33, +4.00]$ & 0.888 & 1.000 \\
AIME 2026 ($N{=}30$) & 1 & 2 & $[-6.67, +16.67]$ & 1.000 & 1.000 \\
MATH-500 ($N{=}500$) & 3 & 4 & $[-0.80, +1.20]$ & 1.000 & 1.000 \\
GPQA Diamond ($N{=}198$) & 3 & 11 & $[+0.51, +8.08]$ & 0.057 & 0.402 \\
OlympiadBench ($N{=}236$) & 10 & 16 & $[-1.69, +6.78]$ & 0.327 & 1.000 \\
\bottomrule
\end{tabularx}
\caption{Paired comparison between Qwen3.8-27B and Qwen3.8-27B-SVRL across all eight external benchmarks. \emph{Qwen3.8-27B only} and \emph{SVRL only} count discordant items passed exclusively by one model. Confidence intervals use 20,000 paired bootstrap resamples (seed 42); $p$ denotes exact two-sided McNemar $p$-values, and $p_{\mathrm{Holm}}$ denotes Holm-adjusted values across the eight-benchmark family.}
\label{tab:external-stats}
\end{table}
}

\newcommand{\externaltokenstable}{
\begin{table}[t]
\centering\small
\setlength{\tabcolsep}{5pt}
\begin{tabularx}{.94\textwidth}{@{}Xrrrr@{}}
\toprule
\textbf{Benchmark} & \textbf{Paired $N$} & \textbf{Qwen3.8-27B} & \textbf{Qwen3.8-27B-SVRL} & \textbf{Change (\%)} \\
\midrule
BFCL-v4 & 800 & 4,142 & 2,516 & $-$39.3\% \\
$\tau^2$-bench Telecom & 113 & 7,058 & 4,304 & $-$39.0\% \\
IFEval & 541 & 2,358 & 2,404 & +2.0\% \\
IFBench & 300 & 8,506 & 6,766 & $-$20.5\% \\
AIME 2026 & 30 & 16,254 & 12,367 & $-$23.9\% \\
MATH-500 & 500 & 1,942 & 1,611 & $-$17.0\% \\
GPQA Diamond & 198 & 10,561 & 7,821 & $-$25.9\% \\
OlympiadBench & 236 & 13,100 & 9,557 & $-$27.0\% \\
\bottomrule
\end{tabularx}
\caption{Mean output tokens per task for Qwen3.8-27B and Qwen3.8-27B-SVRL (including unsuccessful items and returned reasoning tokens, excluding user-simulator tokens). On $\tau^2$-bench Telecom, 113 tasks retain complete paired trajectories; success counts include the 114th task whose Qwen3.8-27B failure has no recoverable dialogue.}
\label{tab:external-tokens}
\end{table}
}

\newcommand{\innob}{\textsc{InnoBench}}

\hypersetup{
  hidelinks,
  pdfauthor={Shengyao Wang, Siyuan Su, Siyu Song, Mengsong Wu, Yiming Luo, Wenbo Wu, Hao Hao, Aimin Zhou},
  pdftitle={Learning What to Change: State-Verified Reinforcement Learning for Educational Agents}
}

\title{Learning What to Change:\\State-Verified Reinforcement Learning for Educational Agents}
\author{%
  \textbf{Shengyao Wang\textsuperscript{1}\quad Siyuan Su\textsuperscript{1}\quad
    Siyu Song\textsuperscript{1}\quad Mengsong Wu\textsuperscript{1}}\\
  \textbf{Yiming Luo\textsuperscript{1}\quad Wenbo Wu\textsuperscript{1}\quad
    Hao Hao\textsuperscript{1}\quad Aimin Zhou\textsuperscript{1,2}}\\[2pt]
  {\normalfont\normalsize\textsuperscript{1}Shanghai Institute of AI for Education, East China Normal University}\\
  {\normalfont\normalsize\textsuperscript{2}Shanghai Innovation Institute}%
}

\begin{document}
\maketitle
\raggedbottom
\renewcommand{\dbltopfraction}{.92}
\renewcommand{\dblfloatpagefraction}{.88}
\renewcommand{\textfraction}{.06}
\setlength{\textfloatsep}{5pt plus 1pt minus 1pt}
\setlength{\dbltextfloatsep}{5pt plus 1pt minus 1pt}
\setlength{\dblfloatsep}{3.5pt plus 1pt minus 1pt}
\setlength{\floatsep}{4pt plus 1pt minus 1pt}
\setlength{\abovedisplayskip}{4pt plus 1pt minus 1pt}
\setlength{\belowdisplayskip}{4pt plus 1pt minus 1pt}
\setlength{\abovedisplayshortskip}{2.5pt plus 1pt minus 1pt}
\setlength{\belowdisplayshortskip}{2.5pt plus 1pt minus 1pt}

\begin{abstract}
Persistent tutoring agents do more than answer questions: they manage course files, practice records, and learner profiles that govern future lessons.
In such stateful settings, knowing what to teach does not guarantee knowing what to change.
We identify a pervasive failure mode, \emph{content--state decoupling}, where agents deliver pedagogically sound responses while silently corrupting workspace state through unwarranted updates, missed prerequisites, or unintended side effects.
Text-level judges overlook these state errors, whereas isolated tool-call checks break down because whether a write action is valid depends jointly on the learner request, prior state, and execution order.
We propose \textbf{State-Verified Reinforcement Learning}, which supervises online policy optimization directly through verifiable workspace transitions.
Within isolated sandboxes that snapshot state around every tool call, a deterministic verifier audits tool effects, workflow ordering, and native auto-sync side effects, pairing graded partial credit with a hard gate that zeroes episode reward upon any critical violation.
Whenever no critical violation occurs, an evidence-grounded judge scores actual saved artifacts rather than agent claims.
We pair 1,000 verifiable training instances with an independently authored, human-reviewed evaluation benchmark of 60 tasks spanning 11 educational categories and 16 native tools.
After online GRPO training, Qwen3.8-27B raises category-weighted strict success on our benchmark from 51.47\% to 80.74\% (+29.27 points), cutting content-pass-only execution failures from 45.6\% to 13.7\% and required state-check failures from 63 to 1 across 180 trials.
The resulting policy outperforms seven reference baselines by over 20 points while reducing output tokens by 17--39\% across seven of eight external benchmarks without regressing on general reasoning or tool use.
\end{abstract}

\section{Introduction}
\label{sec:intro}

Personal learning assistants have moved beyond single-turn question answering.
Persistent tutoring systems maintain longitudinal student workspaces spanning course materials, exercise banks, learner mastery profiles, event logs, knowledge wikis, session histories, and study reminders.
In such environments, every write operation alters durable state that governs later sessions: when a student returns next week, a mastery record should reflect problems the student actually solved rather than worked examples written by the tutor.
Delivering a clear lesson is therefore only half the task; the agent must also decide what to change---and what to leave untouched.

\begin{figure*}[t]
\centering
\resizebox{.87\textwidth}{!}{\input{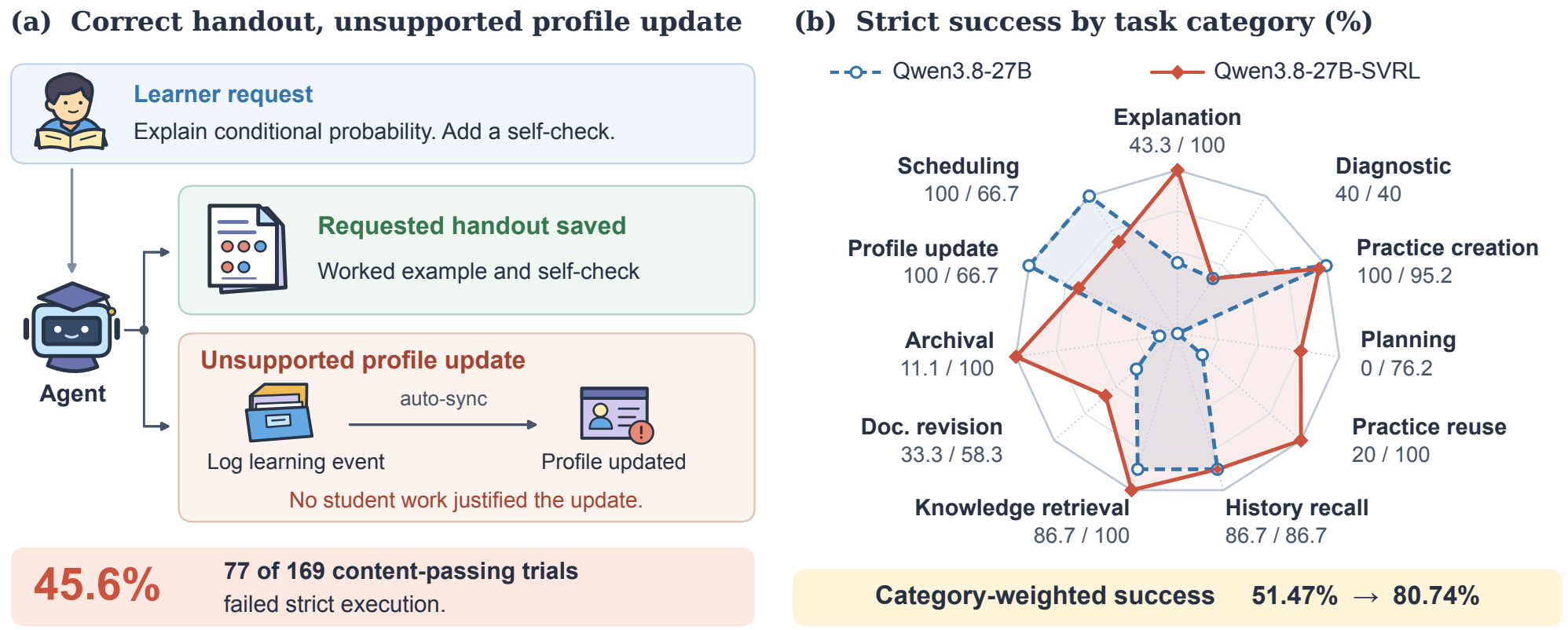}}
\caption{\textbf{(a)} Content--state decoupling in stateful tutoring. On a representative explanation task (Appendix~\ref{app:behavior}), Qwen3.8-27B saves an accurate handout that passes content assessment, yet logs a learning event and triggers an automatic profile update without a student submission. Across 180 Qwen3.8-27B trials, 77 of 169 content passes (45.6\%) fail strict execution. \textbf{(b)} Strict success across all 11 educational task categories. State-Verified RL lifts category-weighted strict success from 51.47\% to 80.74\%, reaching 100\% on four categories and raising clarification and planning from 0\% to 76.2\%.}
\label{fig:motivation}
\end{figure*}

Reinforcement learning (RL) has driven strong gains for tool-using agents in math, coding, and search, where supervision largely reduces to verifying a final answer \citep{jin2025searchr1,feng2026retool,qian2025toolrl}.
Stateful tutoring breaks this assumption.
A tutoring request admits many valid explanations or study plans, yet enforces strict invariants on which workspace records may change and in what order.
Evaluating Qwen3.8-27B across 180 restored workspace trials reveals a pervasive failure mode we term \emph{content--state decoupling} (Figure~\ref{fig:motivation}a).
In a representative explanation task, a student asks for help with conditional probability and a self-check handout.
The model saves a clear, accurate handout that passes content assessment, but then logs a learning event, triggering an automatic learner-profile update even though the student submitted no work.
Across all 180 trials, 169 (93.9\%) pass content assessment, yet 77 of those 169 (45.6\%) fail strict completion because of unauthorized profile edits, skipped prerequisites, or unintended side effects.

Neither response-level judges nor isolated tool-call rules can close this gap.
For example, the same event-logging operation records valid progress after a new submission, corrupts the learner profile during a concept explanation (Figure~\ref{fig:motivation}a), and double-counts evidence when the attempt is already on file.
Whether an action should be rewarded therefore depends jointly on the request $q$, the initial workspace $s_0$, the order of events in trajectory $\tau$, and the final workspace $s_T$.

We address content--state decoupling with \textbf{State-Verified Reinforcement Learning}, grounding policy optimization directly in verifiable workspace transitions.
Each rollout runs in an isolated sandbox restored from $s_0$ with 16 native tools and a scripted learner, capturing workspace state snapshots around every tool call.
Against a private task contract, a deterministic verifier audits target tool effects and workflow ordering while replaying native auto-sync rules to separate authorized runtime updates from critical violations, such as unsupported mastery edits, duplicate attempts, or writes to protected files.
When no critical violation occurs, an independent judge scores pedagogical quality directly from saved artifacts and tool returns rather than agent claims, requiring explicit evidence citations.
We combine these signals into a hard-gated terminal reward for group-relative policy optimization (GRPO): graded execution and content scores preserve partial credit across valid rollouts, while any critical violation zeroes episode reward so fluent prose never offsets state corruption.

To support both training and evaluation over the same 16-tool runtime, we pair 1,000 verifiable training instances with \innob{}, a human-reviewed benchmark of 60 independently authored tasks spanning 11 educational categories.
Online GRPO training from Qwen3.8-27B raises category-weighted strict success from 51.47\% to 80.74\% (+29.27 points), reduces required state-check failures from 63 to 1 across 180 trials, and outperforms seven reference baselines by over 20 points.
Category-level results (Figure~\ref{fig:motivation}b) show the largest gains where state preservation and execution order are decisive: explanation rises from 13/30 to 30/30, practice reuse from 3/15 to 15/15, and clarification and planning from 0/21 to 16/21, while selective, evidence-conditioned profile updates in diagnostic feedback (12/30) remain more challenging than restraint.
Across eight unseen external benchmarks, the trained policy maintains general reasoning and tool-use performance while reducing output tokens by 17--39\% on seven suites.

Our main contributions are:
\begin{itemize}\itemsep1pt\parskip0pt
\item We identify and quantify \emph{content--state decoupling} in stateful tutoring agents: across evaluated baselines, strong reasoning and tool-use capabilities still leave strict completion at or below 60.20\%, and 45.6\% of content-passing Qwen3.8-27B trials fail state or workflow verification.
\item We propose \textbf{State-Verified Reinforcement Learning}, combining workspace state snapshots, native auto-sync replay, a hard-gated terminal reward, and evidence-grounded content judging to supervise online GRPO.
\item We construct 1,000 verifiable training instances alongside \innob{}, a human-reviewed benchmark of 60 tasks over 16 native tools; online GRPO training raises category-weighted strict success by +29.27 points, cuts required state-check failures from 63 to 1, and cuts output tokens by 17--39\% on seven of eight external benchmarks without reasoning or tool-use regression.
\end{itemize}

\section{Related Work}
\label{sec:related}

\paragraph{RL for Tool-Using Agents.}
Interleaving reasoning with tool calls enables agents to adapt to live environment observations \citep{yao2022react}.
Search-R1, ReTool, and ToolRL train search and tool-use policies from interaction outcomes \citep{jin2025searchr1,feng2026retool,qian2025toolrl}, while GiGPO extends credit assignment across both episode and step levels \citep{feng2025gigpo}.
For open-ended tasks without exact ground truth, ArenaRL \citep{zhang2026arenarl} replaces noisy pointwise LLM scores with process-aware pairwise tournaments.
Stateful tutoring raises a complementary supervision problem: even when pedagogical quality can be judged reliably, persistent workspaces enforce hard state invariants conditioned on $s_0$ and prior tool returns.
We therefore couple deterministic state verification with GRPO \citep{shao2024deepseekmath} to supervise both teaching quality and state integrity.

\paragraph{Executable and Stateful Benchmarks.}
AppWorld checks programmatic state transitions and unintended side effects across applications \citep{trivedi2024appworld}, while $\tau$-bench and $\tau^2$-bench evaluate policy compliance and multi-trial consistency in interactive workflows \citep{yao2025taubench,barres2026tau2}.
Agentic Rubrics further ties executable verification criteria to heterogeneous execution evidence \citep{chen2026agentic}.
Educational workspaces place open-ended teaching artifacts and strict state invariants inside the same interaction.
Building on this state-based approach, we pair \innob{} with 1,000 verifiable training instances, bringing state verification from evaluation into online RL.

\paragraph{LLM Tutoring Agents.}
MathDial and MathTutorBench evaluate pedagogical dialogue quality \citep{macina2023mathdial,macina2025mathtutor}, while SocraticLM and TutorRL optimize multi-turn teaching guidance within single problem-solving sessions \citep{liu2024socraticlm,dinucu2025tutorrl}.
Unlike stateless tutoring dialogues, personal learning agents such as Inno Agent maintain and mutate persistent layered memory across sessions \citep{hao2026innoagent}, requiring joint optimization of pedagogical quality and workspace state integrity.

\section{Problem Formulation}
\label{sec:problem}

\subsection{Stateful Educational Trajectories}
\label{sec:formulation}

We formulate each persistent tutoring task as a tuple $(q, s_0, \mathcal{C})$, comprising a learner request $q$, a restorable initial workspace $s_0$, and a private verification contract $\mathcal{C}$.
Each workspace state $s_t$ maintains five persistent stores: workspace files, an L1 learner profile (tracking concept mastery, confidence, stability, misconceptions, and preferences), an append-only learning-event log, retrieval memory (an L2 knowledge wiki and L3 session history), and scheduled reminders.
Interacting over a 16-tool interface $\mathcal{T}$ and a learner simulator, policy $\pi_\theta$ generates a multi-turn trajectory
\begin{equation}
\tau = [z_1, a_1, o_1, \dots, z_K, a_K, o_K, y],
\label{eq:traj}
\end{equation}
where $z_k$, $a_k$, $o_k$, and $y$ denote reasoning, actions (tool calls or clarification questions), returned observations, and the final response, respectively.
Across tool executions, these actions induce a sequence of workspace states $s_0 \to s_1 \to \dots \to s_T$, terminating in final workspace $s_T$.
Rather than prescribing a single fixed tool sequence, $\mathcal{C}$ specifies target state outcomes and invariants, allowing distinct valid explanations, document structures, or error-recovery steps to satisfy the task.

We define \emph{strict task success} as the conjunction of pedagogical quality and operational compliance:
\begin{equation}
S(\tau) = B(\tau) \cdot V(\tau),
\label{eq:strict}
\end{equation}
where $B(\tau) \in \{0,1\}$ indicates that the episode passes independent content assessment, and $V(\tau) \in \{0,1\}$ indicates that it terminates normally within resource budgets, satisfies all mandatory checks, and incurs zero critical violations.

\subsection{Why Content- and Call-Only Rewards Fail}
\label{sec:pathology}

Base-policy rollouts expose three failure modes of conventional rewards in persistent workspaces:

\begin{enumerate}\itemsep2pt\parskip0pt
\item \textbf{Content scores mask state corruption.} The base policy satisfies $B(\tau)=1$ on 169/180 trials (93.9\%) yet reaches $S(\tau)=1$ on only 92 (51.1\%), with 63 failing a required state check and 84 triggering a side-effect violation. Because a text-only judge already scores nearly 94\% of rollouts near ceiling, group-relative updates lose discriminative signal against silent record corruption.
\item \textbf{Action validity depends on $(q, s_0, \tau)$.} Isolated tool-call rules fail because the same write operation changes validity with context and timing. Logging a learning event is required after a new submission, unauthorized during an explanation (where native sync mutates mastery), and a duplicate violation if the same attempt is already recorded in $s_0$. Likewise, saving a study plan before eliciting goals and daily budgets violates workflow order even if the file reads well.
\item \textbf{Partial progress and state corruption are asymmetric.} A study plan that satisfies two of three requirements still provides useful learning signal over an empty rollout, whereas silently inflating mastery, duplicating an attempt, or overwriting a protected file corrupts durable learner state regardless of clarity.
\end{enumerate}

A reliable reward must therefore verify three properties across $(s_0, \tau, s_T)$: whether the requested artifacts and updates were realized, whether prerequisites were gathered \emph{before} committing writes, and whether protected records remained untouched.

\section{State-Verified Reinforcement Learning}
\label{sec:method}

\begin{figure*}[t]
\centering
\resizebox{.90\textwidth}{!}{\input{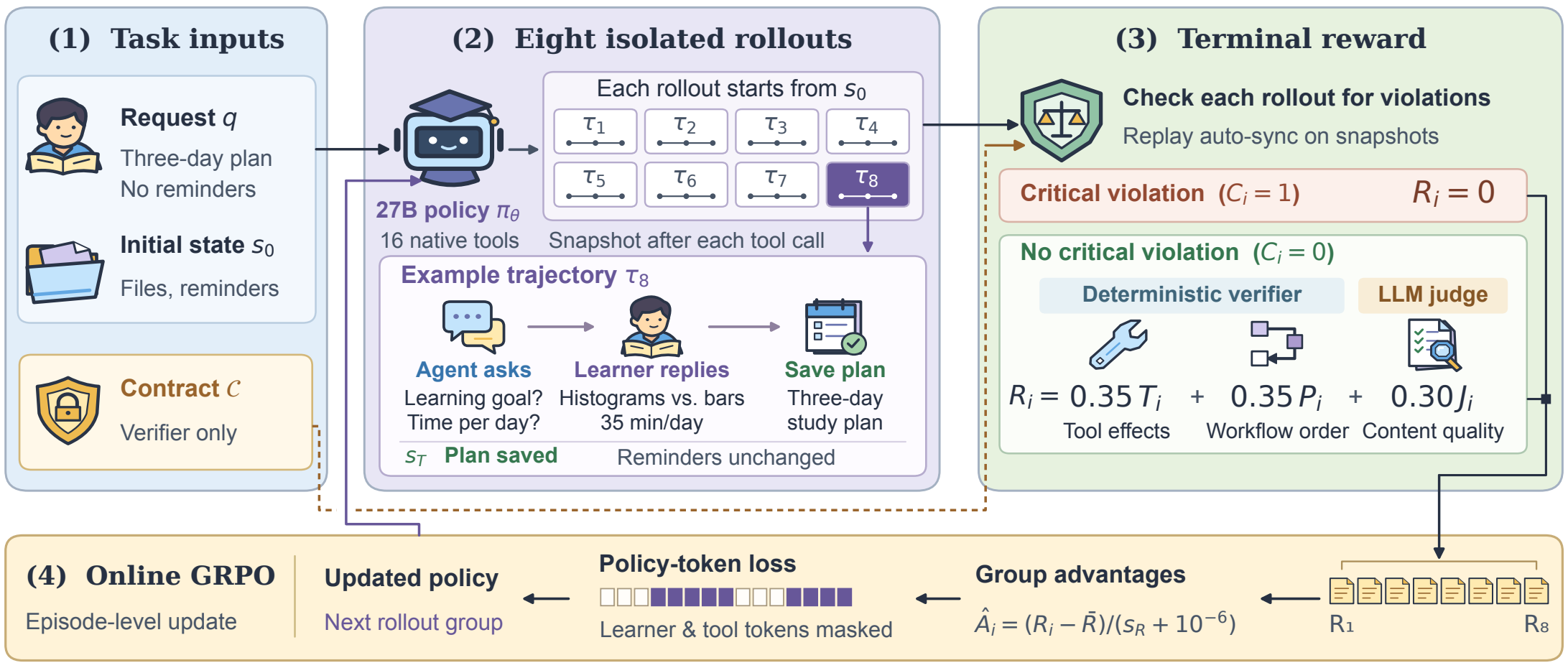}}
\caption{\textbf{Overview of State-Verified Reinforcement Learning.} \textbf{(1)} Each task pairs a learner request $q$ and an initial workspace $s_0$ with a private verification contract $\mathcal{C}$ specifying target effects, workflow order, and protected state. \textbf{(2)} Policy $\pi_\theta$ samples $G{=}8$ isolated rollouts from identical copies of $s_0$ over 16 native tools, recording workspace state snapshots ($s_0 \to \dots \to s_T$) around every tool call. \textbf{(3)} A deterministic verifier scores tool effects $T_i$ and workflow order $P_i$, and detects critical violations $C_i$ via native auto-sync replay; when $C_i{=}0$, an evidence-grounded judge inspects saved artifacts and tool returns to compute $J_i$, yielding the hard-gated terminal reward $R_i$. \textbf{(4)} Online GRPO normalizes rewards within each group and updates only policy-generated tokens while masking tool observations and learner turns.}
\label{fig:framework}
\end{figure*}

Figure~\ref{fig:framework} provides an overview of our training pipeline.
To supervise both pedagogical quality and workspace integrity, State-Verified RL integrates four components: isolated sandbox rollouts with workspace state snapshots, deterministic state and workflow verification with native auto-sync replay, a hard-gated terminal reward with evidence-grounded judging, and online GRPO.

\subsection{Isolated Sandboxed Rollouts and Snapshots}
\label{sec:rollout}

For each task, we spawn $G{=}8$ isolated rollouts from identical copies of $s_0$ inside the workspace runtime, preserving deployment tool schemas and memory-update behavior.
The policy interacts with 16 native tools across five functional groups: file I/O (3), learner memory and profiles (5), L2 wiki and L3 history retrieval (3), learner clarification (1), and reminder management (4); full tool specifications appear in Appendix~\ref{app:implementation}.
Within a four-turn interaction budget, a scripted learner responds to both structured and natural-language clarification requests and issues fixed follow-ups without revealing hidden contract criteria.
Before execution, after every tool call, and upon termination, the sandbox records state snapshots of all persistent workspace stores alongside the ordered event trace, capturing even transient mutations that are later overwritten.

\subsection{State, Workflow, and Auto-Sync Verification}
\label{sec:verifier}

Against the private verification contract $\mathcal{C}$, a deterministic verifier computes three structured signals from the recorded snapshots and event trace without prescribing a single rigid tool sequence:

\paragraph{Tool-Effect Score $T(\tau) \in [0,1]$.}
$T(\tau)$ measures the equal-weighted pass rate across contract-specified operational goals, including reading required source files (merging segmented reads), writing valid target artifacts, linking existing exercises, archiving into the L2 wiki with provenance, updating requested preference fields, and applying exact reminder edits.
Header-only reads, empty writes, and unverified agent claims receive no credit.

\paragraph{Workflow-Order Score $P(\tau) \in [0,1]$.}
$P(\tau)$ measures the equal-weighted pass rate across temporal and structural invariants: reading source materials before writing or archiving, eliciting missing learner constraints \emph{before} drafting a study plan, inspecting history before logging an attempt, enforcing unique attempt records, validating saved artifacts, and terminating normally within budget.
Policies may recover from failed intermediate tool calls and still achieve $P(\tau){=}1$.

\paragraph{Native Auto-Sync Replay and Critical Gate $C(\tau) \in \{0,1\}$.}
In the native runtime, logging a learning event automatically updates system-managed profile fields such as evidence links and timestamps.
Directly editing the learner profile before logging the same event can double-count evidence, whereas authorized automatic updates triggered by a valid event should not be penalized.
Our verifier therefore replays the native auto-sync rules over the captured snapshots to reconstruct expected system-managed states, isolating policy-induced edits from runtime synchronization.
The binary gate triggers ($C(\tau){=}1$) if any snapshot reveals: (1) a modification to a protected workspace file, (2) an unauthorized L2 wiki or reminder change, (3) an unrequested learning event or duplicate attempt record, (4) an explicit, unsupported update to mastery, confidence, or stability, or (5) an unauthorized profile or preference edit.

\subsection{Evidence-Grounded Judge and Hard-Gated Reward}
\label{sec:reward}

When $C(\tau){=}0$, an independent LLM judge evaluates pedagogical quality without access to verifier outputs.
Rather than relying on what the agent claims to have done, the judge inspects the conversation alongside the saved workspace and L2 wiki artifacts, final learner-profile, event-log, and reminder states, and tool returns, each indexed with a unique evidence identifier.
It scores correctness ($w_1{=}0.50$), completeness ($w_2{=}0.25$), grounding ($w_3{=}0.15$), and clarity ($w_4{=}0.10$) on a 0--4 scale while citing valid identifiers from this evidence bundle (invalid citations trigger up to three retries), yielding $J(\tau) = \frac{1}{4}\sum_{m=1}^{4} w_m\,\mathrm{score}_m(\tau) \in [0,1]$.

We combine deterministic verification and content assessment into a hard-gated terminal reward:
\begin{equation}
\begin{split}
R(\tau) = (1 - C(\tau))\bigl[&0.35\,T(\tau) + 0.35\,P(\tau)\\
                             &+ 0.30\,J(\tau)\bigr].
\end{split}
\label{eq:reward}
\end{equation}
When $C(\tau){=}0$, the 70/30 balance between graded execution checks ($0.35\,T(\tau){+}0.35\,P(\tau)$) and content quality ($0.30\,J(\tau)$) preserves reward discrimination across partially successful rollouts.
When $C(\tau){=}1$, the gate zeroes $R(\tau)$ and bypasses content evaluation, ensuring fluent responses never offset state corruption while avoiding unnecessary judge calls.
Episodes affected by runtime or snapshot-capture errors are excluded from policy updates rather than assigned fallback scores.

\tasksuitetable

\subsection{Online Group-Relative Policy Optimization}
\label{sec:grpo}

We optimize policy $\pi_\theta$ directly from the base model using synchronous online GRPO \citep{shao2024deepseekmath} without an SFT warm-up stage.
For each task, we sample $G{=}8$ rollouts from identical copies of $s_0$ and compute a single group-normalized advantage per episode, independent of turn count:
\begin{equation}
\widehat{A}_i = \begin{cases}
0, & s_R = 0, \\
(R_i - \bar{R}) / (s_R + 10^{-6}), & s_R > 0,
\end{cases}
\label{eq:grpo}
\end{equation}
where $\bar{R}$ and $s_R$ denote the group mean and sample standard deviation of $\{R_i\}_{i=1}^G$.
Every policy-generated token in $\tau_i$ receives advantage $\widehat{A}_i$, whereas tool observations and learner turns are masked from the loss (Figure~\ref{fig:framework}, Stage~4).
Each optimization step processes four task groups (32 episodes) at a learning rate of $10^{-6}$ for 250 updates, with both KL regularization and entropy coefficients set to zero (Appendix~\ref{app:implementation}).

\section{Training Suite and InnoBench}
\label{sec:benchmark}

We pair 1,000 verifiable training instances with \innob{}, a human-authored benchmark of 60 tasks operating over the same 16-tool workspace runtime (Table~\ref{tab:suite}).

\paragraph{Verifiable Training Suite (1,000 Instances).}
The training suite comprises 1,000 tasks organized into 396 source groups across 36 topics in mathematics, programming, science, and study planning.
Each instance separates public inputs ($q, s_0$) and scripted learner facts from private verification checks and rule-derived reference answers.
To train both multi-turn adaptability and restraint, 320 instances include a fixed learner follow-up, while clarification tasks mix underspecified requests with fully specified prompts that penalize unnecessary questions.

\paragraph{\innob{} (60 Tasks).}
\innob{} consists of 60 independently authored tasks matching the 11-category training distribution, spanning mathematics (20), programming (16), science (15), and study planning (9), with a worked example in Appendix~\ref{app:categories}.
Twenty tasks include fixed learner follow-ups, and five of the seven clarification tasks withhold essential constraints upfront.
Three human reviewers inspected all 60 tasks and audited model execution traces.
We evaluate the final checkpoint after the fixed 250-update schedule on \innob{} alongside eight unseen external benchmarks.

\paragraph{Evaluation Protocol.}
Each model runs three trials per task from restored copies of $s_0$ under uniform execution limits (16,384 generated tokens, 24 turns, 30 tool calls, and a 300-second timeout).
Strict task success requires $S(\tau){=}1$ (Equation~\ref{eq:strict}), using DeepSeek-V4.1-Flash-1017 as the independent content judge.
Our primary metric is category-weighted strict success:
\begin{equation}
\mathrm{Score} = \sum_{c=1}^{11} w_c \cdot \frac{1}{|D_c|}\sum_{x\in D_c}\bar{S}(x),
\label{eq:weighted_score}
\end{equation}
where $w_c$ denotes the training proportion of category $c$, $D_c$ its task subset in \innob{}, and $\bar{S}(x)$ the three-trial mean strict success on task $x$.
We additionally report unweighted trial passes (out of 180) and All-3 consistency (tasks passing all three trials, out of 60).

\section{Experiments}
\label{sec:experiments}

\allmodelstable

\subsection{Experimental Setup}
\label{sec:setup}

We compare Qwen3.8-27B against Qwen3.8-27B-SVRL (trained with State-Verified RL), InnoSpark 3.1 (a prior educational model), and five open-weight reference models: GLM-4.7-Flash, Muse Glimmer 30B High, Gemma 4 26B-A4B Thinking, GPT-OSS-20B High, and OmniEdu-27B \citep{liang2026omniedu}.

All \innob{} evaluations use deterministic decoding (temperature 0, top-$p$ 1) under a unified sandbox runtime, with DeepSeek-V4.1-Flash-1017 serving as the content judge (reasoning mode enabled, 32,768-token budget).
We estimate paired 95\% confidence intervals between Qwen3.8-27B and Qwen3.8-27B-SVRL via 10,000 stratified within-category bootstrap resamples, keeping each task's three trials grouped.
To assess out-of-domain generalization and non-regression, we evaluate all eight models across eight external benchmarks: AIME 2026, MATH-500 \citep{lightman2024verify}, GPQA Diamond \citep{rein2024gpqa}, OlympiadBench \citep{he2024olympiad}, IFEval \citep{zhou2023ifeval}, IFBench \citep{pyatkin2025ifbench}, BFCL-v4 \citep{patil2025bfcl}, and $\tau^2$-bench Telecom \citep{barres2026tau2}.
Paired differences between Qwen3.8-27B and Qwen3.8-27B-SVRL on these external suites are tested using exact two-sided McNemar tests with Holm correction (Appendix~\ref{app:external}).

\subsection{Main Results on InnoBench}
\label{sec:mainres}

\mainresultstable

Table~\ref{tab:allmodels} compares all eight models across nine suites; Table~\ref{tab:main} breaks down Qwen3.8-27B vs.\ Qwen3.8-27B-SVRL on \innob{}.

\paragraph{Closing Content--State Decoupling.}
State-Verified RL lifts category-weighted strict success from 51.47\% to 80.74\% (+29.27 percentage points; paired 95\% CI $[21.80, 36.64]$), increasing strict trial passes from 92/180 to 145/180 and All-3 consistency from 23/60 to 42/60 while reducing zero-pass tasks (0/3) from 21/60 to 6/60.
Meanwhile, content pass rates remain virtually unchanged: Qwen3.8-27B passes 169/180 trials (93.9\%), and Qwen3.8-27B-SVRL passes 168/179 judged trials (93.9\%, as the single unjudged trial already failed an objective diagnostic check).
Nearly the entire 29-point gain comes from cutting \emph{content-pass-only execution failures} from 77/169 (45.6\%) to 23/168 (13.7\%): Qwen3.8-27B knew what to teach; State-Verified RL teaches what to change.

\paragraph{Cross-Model Positioning.}
Across all eight models (Table~\ref{tab:allmodels}), Qwen3.8-27B-SVRL (80.74\%) outperforms the strongest external baseline, Gemma 4 26B-A4B Thinking (60.20\%), by over 20 points, followed by Muse Glimmer 30B High (55.20\%), InnoSpark 3.1 (53.07\%), Qwen3.8-27B (51.47\%), OmniEdu-27B (51.18\%), GPT-OSS-20B High (43.58\%), and GLM-4.7-Flash (39.21\%).
Neither strong tool use nor domain reasoning alone ensures stateful compliance: GLM-4.7-Flash leads both BFCL-v4 (56.38\%) and $\tau^2$-bench Telecom (96.49\%) yet scores 39.21\% on \innob{}, while OmniEdu-27B leads OlympiadBench (50.42\%) yet matches Qwen3.8-27B on \innob{} (51.18\%).

\subsection{Behavioral and Error Analysis}
\label{sec:analysis}

\begin{figure*}[t]
\centering
\resizebox{.84\textwidth}{!}{\input{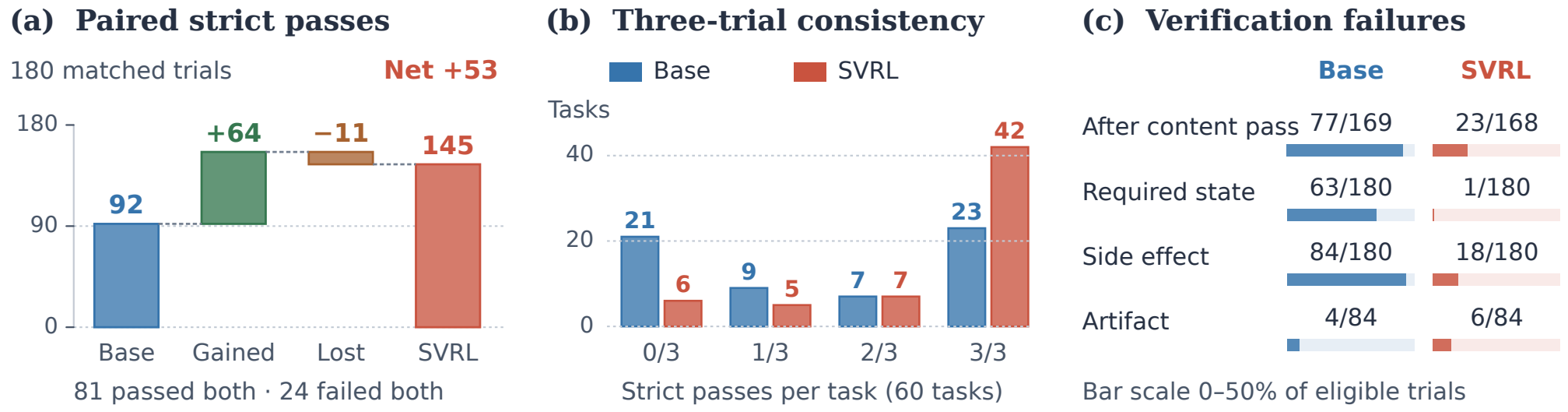}}
\caption{\textbf{Behavioral shift after State-Verified RL on \innob{} (60 tasks $\times$ 3 trials; Base: Qwen3.8-27B, SVRL: Qwen3.8-27B-SVRL).} \textbf{(a)} Matched-trial comparison: 64 Qwen3.8-27B failures flip to strict passes under Qwen3.8-27B-SVRL against 11 regressions (net +53). \textbf{(b)} Three-trial consistency: All-3 tasks rise from 23 to 42 while 0/3 failures fall from 21 to 6. \textbf{(c)} Failure rates by verification check: required state-check failures drop from 63/180 to 1/180, and side-effect failures drop from 84/180 to 18/180.}
\label{fig:behavior}
\end{figure*}

Figure~\ref{fig:behavior} decomposes the 29.27-point gain.
Matched by task and trial (Figure~\ref{fig:behavior}a), 64 Qwen3.8-27B failures turn into strict passes under Qwen3.8-27B-SVRL against 11 regressions, while 81 trials pass and 24 fail under both models.
Required state-check failures drop sharply from 63/180 (35.0\%) to 1/180 (0.6\%), and side-effect failures fall from 84/180 (46.7\%) to 18/180 (10.0\%) (Figure~\ref{fig:behavior}c), whereas artifact-check failures change only marginally (from 4/84 to 6/84).

Across the 11 task categories (Figure~\ref{fig:motivation}b; Table~\ref{tab:categories}, Appendix~\ref{app:categories}), four clear patterns emerge:
(i) \emph{State preservation and non-duplication reach 100\%}: Explanation (13/30 to 30/30), Practice Reuse (3/15 to 15/15), Knowledge Archival (1/9 to 9/9), and Knowledge Retrieval (13/15 to 15/15).
(ii) \emph{Prerequisite ordering recovers from zero}: Clarification and Planning rises from 0/21 to 16/21 (+76.2 percentage points), and Document Revision improves from 4/12 to 7/12.
(iii) \emph{Small write-only categories show mild caution}: Profile Update and Scheduling (six trials each) dip slightly from 6/6 to 4/6.
(iv) \emph{Diagnostic feedback stays flat at 12/30} (Section~\ref{sec:casestudy}).

\subsection{Training Dynamics, Conciseness, and Generalization}
\label{sec:dynamics}

\begin{figure*}[t]
\centering
\resizebox{.84\textwidth}{!}{\input{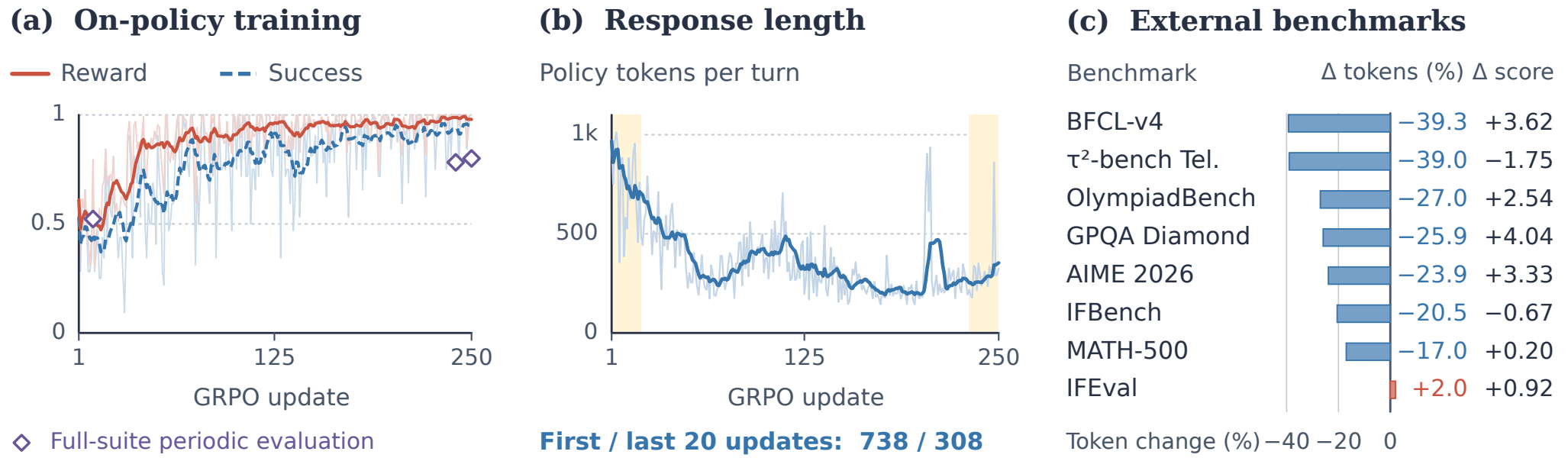}}
\caption{\textbf{Training dynamics across 250 online GRPO updates and out-of-domain generalization.} \textbf{(a)} On-policy reward and training success (raw values and 10-update moving averages) climb steadily from Qwen3.8-27B; diamonds mark the three completed full-suite periodic evaluations without interpolation. \textbf{(b)} Mean per-turn policy response length decreases by 58.3\% (from 738 to 308 tokens). \textbf{(c)} Output token change (\%) and paired accuracy difference ($\Delta$ score, in percentage points) from Qwen3.8-27B to Qwen3.8-27B-SVRL across eight unseen benchmarks: Qwen3.8-27B-SVRL reduces output tokens by 17--39\% on seven suites without capability regression.}
\label{fig:dynamics}
\end{figure*}

Figure~\ref{fig:dynamics}a tracks training progression across all 250 online GRPO updates from Qwen3.8-27B.
With graded partial credit sustaining intra-group variance from the first update, mean on-policy reward climbs from 0.557 (first 20 updates) to 0.984 (last 20), and training success rises from 44.53\% to 94.53\%.
The three scheduled checkpoints that completed full 180-trial periodic evaluation (under non-reasoning judging) follow the same curve without interpolation: 52.26\% (step 10), 78.24\% (240), and 80.03\% (250; Table~\ref{tab:training-checkpoints}, Appendix~\ref{app:implementation}).

State-Verified RL also substantially shortens policy responses: mean per-turn response length falls by 58.3\% (from 738 to 308 tokens; Figure~\ref{fig:dynamics}b).
This conciseness transfers out of domain (Figure~\ref{fig:dynamics}c; Table~\ref{tab:external-tokens}): Qwen3.8-27B-SVRL reduces mean output tokens on seven of the eight unseen benchmarks, including $-39.3\%$ on BFCL-v4 (from 4,142 to 2,516), $-39.0\%$ on $\tau^2$-bench Telecom (from 7,058 to 4,304), $-27.0\%$ on OlympiadBench, $-25.9\%$ on GPQA Diamond, $-23.9\%$ on AIME 2026, $-20.5\%$ on IFBench, and $-17.0\%$ on MATH-500, with only a $+2.0\%$ change on IFEval.

Crucially, this conciseness comes without capability regression (Table~\ref{tab:allmodels}): Qwen3.8-27B-SVRL ranks first (or tied) on AIME 2026, MATH-500, and GPQA Diamond (second on OlympiadBench behind OmniEdu-27B) and improves across all four BFCL-v4 subsets (50.88\% to 54.50\%, unadjusted $p{=}0.021$; Table~\ref{tab:bfcl-subsets}), while matching Qwen3.8-27B on $\tau^2$-bench Telecom (100/114 vs.\ 102/114, $p{=}0.815$), IFEval, and IFBench (smallest Holm-adjusted $p{=}0.172$; Table~\ref{tab:external-stats}, Appendix~\ref{app:external}).

\subsection{Case Study: Restraint vs.\ Selective Diagnosis}
\label{sec:casestudy}

To understand why Qwen3.8-27B-SVRL reaches 30/30 on Explanation yet remains at 12/30 on Diagnostic Feedback, we examine paired trajectories on a representative diagnostic task (Appendix~\ref{app:behavior}), in which a student revisits a Python submission that miscounted duplicate elements by using a set.
Qwen3.8-27B-SVRL reads the workspace notes, accurately explains why a Python set discards multiplicities, and proceeds to log a learning event and directly update the learner profile.
However, because it modifies the profile without first inspecting prior learner context, it triggers profile-mutation and learner-state violations.
On the identical trial, InnoSpark 3.1 retrieves prior learner context before logging the event and updating the profile, passing every verification check.

This contrast highlights a structural limitation of episode-level RL.
Explanation and reuse tasks reward \emph{blanket restraint}, which a terminal hard gate teaches rapidly (reducing required state-check failures from 63 to 1), whereas follow-up diagnosis requires \emph{selective, context-conditioned edits}.
Broadcasting a single episode advantage $\widehat{A}_i$ uniformly across all turns makes separating grounded inspect-then-update sequences from ungrounded edits harder than suppressing writes.

\section{Conclusion}
\label{sec:conclusion}

In persistent tutoring workspaces, delivering clear instruction and maintaining trustworthy learner records are distinct capabilities.
We identified and quantified \emph{content--state decoupling} in stateful tutoring agents---where 45.6\% of content-passing trials from Qwen3.8-27B still fail state or workflow verification---and demonstrated that State-Verified RL substantially closes this gap.
Through online GRPO supervised by workspace state snapshots, workflow verification, native auto-sync replay, a hard-gated terminal reward, and evidence-grounded content judging, Qwen3.8-27B-SVRL raises category-weighted strict success on \innob{} from 51.47\% to 80.74\%, reduces required state-check failures from 63 to 1 across 180 trials, and cuts output tokens by 17--39\% on seven of eight external benchmarks without regressing on general reasoning or tool use.
More broadly, our findings show that verifiable state transitions provide a scalable supervision signal for building trustworthy, persistent agents beyond single-turn interaction.

\clearpage
\section*{Limitations}

While State-Verified RL substantially improves stateful tutoring reliability, our study has several scope boundaries that motivate future work.
First, our experiments focus on training the full State-Verified RL objective on a single 27B backbone; systematic ablations across individual reward components ($T$, $P$, $J$, and $C$) and evaluations across broader model scales remain valuable extensions.
Second, scripted learners enable controlled, reproducible cross-model comparisons from identical initial workspaces, whereas measuring long-term learning gains requires longitudinal studies with real students.
Finally, while episode-level GRPO advantages rapidly instill state preservation and workflow ordering, tasks requiring selective multi-field profile updates (such as diagnostic feedback) may benefit further from finer-grained, turn-level credit assignment.

\section*{Ethics and Reproducibility}

All training instances and \innob{} tasks are constructed with synthetic learner personas, synthetic requests, and restorable sandboxed workspaces, containing no real student data or personally identifiable information.
To ensure auditability and reproducibility, each evaluation run logs task identifiers, random seeds, trial indices, tool-call traces, workspace state snapshots, and content-judge outputs; decoding settings, training hyperparameters, and full evaluation details are documented in Appendix~\ref{app:implementation}.
Generative AI tools were used solely to assist with literature discovery, manuscript editing, and figure preparation; all experimental results and statistical tests were manually verified against the raw evaluation logs.


\clearpage
\onecolumn
\appendix
\let\appendixtable\table
\let\endappendixtable\endtable
\renewenvironment{table}[1][]{\appendixtable[!htbp]}{\endappendixtable}
\renewenvironment{table*}[1][]{\appendixtable[!htbp]}{\endappendixtable}
\setlength{\floatsep}{10pt}
\setlength{\textfloatsep}{10pt}
\setlength{\intextsep}{9pt}

\section{Task Coverage and a Worked Planning Example}
\label{app:categories}

The verifiable training suite contains 1,000 instances drawn from 396 source groups across 36 topics, including 320 instances with a fixed learner follow-up.
The 60 \innob{} tasks span mathematics (20), programming (16), science (15), and study planning (9), with 20 tasks including a fixed learner follow-up.
Among the seven clarification and planning tasks, five withhold essential constraints in the initial prompt and two provide all constraints upfront.
Table~\ref{tab:categories} reports the exact category weights and per-category strict pass counts for Qwen3.8-27B and Qwen3.8-27B-SVRL; categories containing two tasks contribute six trials per model.

\categoryresultstable

\paragraph{Worked Planning Example.}
A learner asks for a three-day review plan on statistical charts, requests that the agent save the plan to a file, and explicitly prohibits creating reminders.
The initial prompt omits both the specific learning goal and the daily time budget.
When prompted via either a structured clarification request or a direct natural-language question, the scripted learner specifies that they want to distinguish histograms from bar charts and have 35 minutes per day.
The deterministic verifier checks that: (i) the agent elicits both missing constraints \emph{before} saving the plan, (ii) the saved artifact exists at the target path, (iii) scheduled reminders remain unchanged from $s_0$, and (iv) the planned review is not prematurely logged as completed learning evidence.
The content judge independently evaluates whether the saved file addresses the clarified goal and fits within the 35-minute daily budget.

\section{Paired Case Studies}
\label{app:behavior}

\paragraph{Case 1: Explaining Conditional Probability.}
Figure~\ref{fig:motivation}a summarizes a representative Qwen3.8-27B trajectory on an explanation task.
The model reads the instructional materials twice, saves an accurate worked example and self-check handout, and then logs a learning event for concept explanation.
The write operation succeeds and triggers native automatic synchronization of the learner profile.
Although the content judge awards full credit to the handout, the deterministic verifier flags the resulting knowledge-state mutation because no student submission justified updating the profile.

\paragraph{Case 2: Revisiting a Programming Diagnosis.}
In a representative diagnostic follow-up task, the learner revisits a previously submitted solution involving Python sets that miscounted duplicate elements.
Qwen3.8-27B-SVRL reads the course material, accurately explains why a Python set discards duplicate element frequencies, and proceeds to log a learning event and directly update the learner profile.
The verifier flags both profile-mutation and learner-state violations because the edit was performed without consulting prior learner records.
On the exact same task and trial, InnoSpark 3.1 retrieves prior learner context before logging the event and updating the profile, passing all required checks.
Both trajectories invoke the same two write operations; task success depends on grounding those updates in prior workspace state.

\section{Training and Evaluation Configuration}
\label{app:implementation}

\paragraph{RL Training Setup.}
Online GRPO training initializes from Qwen3.8-27B across 8 GPUs (tensor parallelism 4, pipeline parallelism 2, data parallelism 1, with CPU optimizer offloading) using a learning rate of $10^{-6}$, random seed 17, and zero KL-regularization and entropy coefficients.
Each of the 250 updates samples four task groups of $G{=}8$ episodes (32 trajectories per step).
Training rollouts allow a 32,768-token context window, up to 16,384 generated tokens, 40 model calls, and 32 tool calls, with the scripted learner supporting up to four interaction turns.
All training rollouts and \innob{} evaluations expose the same 16 native workspace tools: file I/O (\texttt{read}, \texttt{write}, \texttt{edit}), learner memory and profiles (\texttt{get\_\allowbreak learner\_\allowbreak context}, \texttt{record\_\allowbreak learning\_\allowbreak event}, \texttt{update\_\allowbreak learner\_\allowbreak profile}, \texttt{patch\_\allowbreak learner\_\allowbreak profile}, \texttt{review\_\allowbreak learner\_\allowbreak profile}), knowledge and history retrieval (\texttt{l2\_\allowbreak query}, \texttt{l2\_\allowbreak archive}, \texttt{l3\_\allowbreak recall}), learner clarification (\texttt{ask\_\allowbreak user\_\allowbreak question}), and reminder management (\texttt{create\_\allowbreak scheduled\_\allowbreak job}, \texttt{list\_\allowbreak scheduled\_\allowbreak jobs}, \texttt{update\_\allowbreak scheduled\_\allowbreak job}, \texttt{delete\_\allowbreak scheduled\_\allowbreak job}).
Episodes with unavailable or invalid reward feedback are excluded from policy updates rather than assigned fallback scores.

\paragraph{InnoBench Evaluation Protocol.}
All main-text \innob{} comparisons are conducted under a unified sandboxed evaluation runtime.
Each trial restores the fixed initial workspace $s_0$ and enforces cumulative episode budgets of 16,384 generated tokens, 24 assistant turns, 30 tool calls, and 300 seconds.
All models decode deterministically at temperature 0 and top-$p$ 1 under their native serving configurations.
DeepSeek-V4.1-Flash-1017 serves as the content judge with reasoning mode enabled and a 32,768-token request budget.
All eight evaluated models use native function calling under this unified interface.

Across all eight models, there are 1,440 distinct model--task--trial records with complete strict outcomes.
Twenty-three episodes lack a completed content judgment (six exceeded an execution budget, 13 failed a mandatory objective check, and four timed out), each of which independently constitutes a strict failure.
Reported content pass rates exclude unjudged episodes from the denominator: Qwen3.8-27B passes 169/180 judged trials, and Qwen3.8-27B-SVRL passes 168/179 judged trials (its single unjudged trial failed an objective diagnostic check).

\paragraph{Training Dynamics and Periodic Checkpoints.}
Comparing the first 20 against the last 20 training updates across the 250-step run, mean on-policy episode reward rises from 0.557 to 0.984, logged training success increases from 44.53\% to 94.53\%, and mean per-turn policy response length decreases from 738 to 308 tokens.
Table~\ref{tab:training-checkpoints} reports the three completed full-suite periodic evaluations (180 trials each, scored under non-reasoning judging), whereas the final comparisons in Tables~\ref{tab:allmodels} and~\ref{tab:main} use the unified reasoning-enabled evaluation protocol.

\trainingresultstable

\FloatBarrier
\section{External Benchmark Protocols and Statistical Tests}
\label{app:external}

\paragraph{Benchmark Protocols.}
The six single-response benchmarks evaluate one response per item under a 32,768-token generation budget.
IFEval (541 prompts) and IFBench (300 prompts) use strict prompt-level scoring; GPQA uses the 198-item Diamond split; OlympiadBench evaluates 236 English text physics problems with full context using the official symbolic grader; AIME 2026 (30 items) and MATH-500 (500 items) use exact mathematical equivalence.
Under a fixed generation allowance, failing to terminate within budget yields an empty final answer (for example, GPT-OSS-20B High exhausts the budget on 447/541 IFEval prompts and 239/300 IFBench prompts).

For BFCL-v4, we evaluate 200 tasks from each of four multi-turn subsets (800 tasks in total; Table~\ref{tab:bfcl-subsets}) using official prompts, native checkers, and a 20-step limit.
For $\tau^2$-bench Telecom (114 tasks), all models use native function calling and interact with the same Qwen3.8-27B user simulator under the official environment reward, with limits of 100 steps and 10 tool errors.

\bfclsubsettable

\FloatBarrier
\paragraph{Paired Statistical Tests and Token Usage.}
Each external benchmark evaluates Qwen3.8-27B and Qwen3.8-27B-SVRL on identical items with one attempt per item.
Table~\ref{tab:external-stats} reports discordant counts, unadjusted 95\% percentile intervals from 20,000 paired bootstrap resamples (seed 42), exact two-sided McNemar $p$-values, and Holm-adjusted $p$-values across the eight-benchmark family.
Table~\ref{tab:external-tokens} reports mean output tokens per task for Qwen3.8-27B and Qwen3.8-27B-SVRL (including reasoning tokens when returned by the serving stack and excluding user-simulator tokens).

\externalstatstable
\externaltokenstable
\FloatBarrier

\end{document}